\documentclass[a4paper,twoside]{article}
\usepackage{epsfig}
\usepackage{subfigure}
\usepackage{natbib}
\usepackage{url}
\usepackage{breakurl}

\usepackage[breaklinks]{hyperref}
\usepackage{calc}
\usepackage{amssymb}
\usepackage{amstext}
\usepackage{amsmath}
\usepackage{amsthm}
\usepackage{multicol}
\usepackage{pslatex}
\usepackage{fancyhdr}

\usepackage{SCITEPRESS}     

\begin{document}

\title{Time Stretch Inspired Computational Imaging}
\author{\authorname{Bahram Jalali\sup{1,2,3}, Madhuri Suthar\sup{1}, Mohamad Asghari\sup{1} and Ata Mahjoubfar\sup{1,2}}
\affiliation{\sup{1}Department of Electrical Engineering, University of California Los Angeles, Los Angeles, California, USA}
\affiliation{\sup{2}California NanoSystems Institute, Los Angeles, California, USA}
\affiliation{\sup{3}Department of Bioengineering, University of California Los Angeles, Los Angeles, California, USA}
}



\abstract{We show that dispersive propagation of light followed by phase detection has properties that can be exploited for extracting features from the waveforms. This discovery is spearheading development of a new class of physics-inspired algorithms for feature extraction from digital images with unique properties and superior dynamic range compared to conventional algorithms. In certain cases, these algorithms have the potential to be an energy efficient and scalable substitute to synthetically fashioned computational techniques in practice today.}

\onecolumn \maketitle \normalsize \vfill

\section{\uppercase{Introduction}}
\label{sec:introduction}

\noindent ``Human subtlety will never devise an invention more beautiful, more simple or more direct than does nature''. The elegant quote by Leonardo Da Vinci underscores the important role of nature as a source of inspiration for human ingenuity. Inspirations from nature need not be limited to design of physical machines but should be extended to creation of new computational algorithms. We expect this new paradigm to lead to a new class of algorithms that are direct and energy efficient while providing unprecedented functionality.  

Every day, the world creates nearly 2.5 exabyte ($10^{18}$ bytes) of data. Surprisingly, 90\% of the data present in the world today has been created in the last two years alone highlighting the exponential increase in the amount of digital data (\cite{IBMURL}). Processing this massive data in datacentres accounts for 50-60\% of their electricity budget and a rapidly growing fraction of total electricity consumption. This calls for development of new computing technologies that offer speed, energy efficiency and ease of implementation. Fortunately nature, and in particular optics-inspired algorithms can provide a solution to certain class of problems. 

Recently, optical hardware accelerators have been proposed as a mean to boost the speed and reduce the power consumption of electronics (\cite{jalali2015tailoring}). In particular, it was shown that one can create an analog optical gearbox for matching the time-bandwidth of fast real-time optical data to that of the much slower electronics. This information gearbox can enable real-time processing of ultrafast optical data while reducing the power consumption and improving the sensitivity. 

\begin{figure*}[!h]
  \centering
   {\epsfig{file = 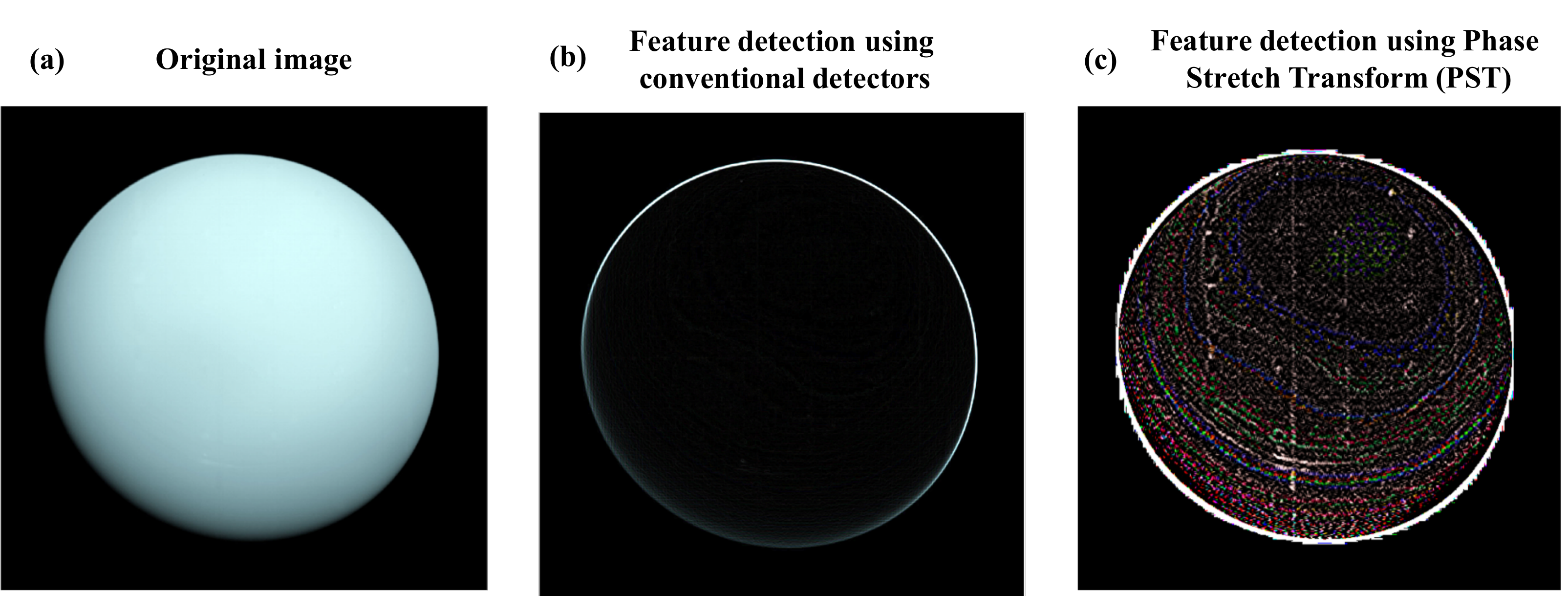, width = 16 cm}}
  \caption{Comparison of feature detection using conventional derivative operator to the case of feature detection using Phase Stretch Operator (PST). The derivative is the fundamental operation used in the popular Canny, Sobel and Prewitt edge detection methods.}
  \label{fig:example1}
 \end{figure*} 
 
At the same time, Phase Stretch Transform (PST) was recently introduced as a new computational approach to signal and image processing  (\citet{asghari2014physics,asghari2015edge}). PST emerged out of research on the Photonic Time Stretch (\citet{bhushan1998time, ng2014demonstration, mahjoubfar2015design, han2003photonic}), a real-time measurement technique that has led to the discovery of optical rogue waves (\cite{solli2007optical}), observation of relativistic electron microstructure (\cite{roussel2014microbunching}), observation of the birth of modelocking (\cite{herink2016resolving}) and record accuracy for label-free cancer cell detection (\cite{chen2016deep}).  PST is a physics-based image processing approach that mimics the propagation of electromagnetic waves through a diffractive medium with engineered dispersive property (refractive index) (\citet{asghari2014physics,asghari2015edge}). The algorithm performs edge detection and feature extraction on both digital images as well as time domain waveforms. It has been used for feature extraction in biomedical images  (\citet{asghari2015edge, suthar2016diagnostic,suthar2016decision}) and Synthetic Aperture Radar (SAR) (\cite{ilioudis2015edge}). It has also been applied to resolution enhancement in super-resolution localization microscopy where it drastically improved the point spread function, reduced the computational time by 400\% and increased the required emitter density by the same amount (\cite{ilovitsh2016phase}). The Phase Stretch Transform algorithm was recently open sourced on GitHub and Matlab Central File Exchange (\cite{GithubURL}),  and has received extraordinary endorsements by the software and image processing community.

PST is a qualitatively new method of image processing that was introduced last year. In this paper, we show for the first time that PST has unique intrinsic properties not offered by the current state-of-the-art algorithms. We show that the new algorithm reveals features invisible to human eye and to conventional algorithms being used today. Below we prove and explain this property using the mathematical formulation of the PST response function.

To demonstrate this point, Figure 1 shows an image of the planet Uranus processed by the conventional edge detection algorithm (derivative based) and by the PST. The derivative method is the underlying function utilized by the popular Canny, Sobel and Prewitt algorithms.  The result clearly shows the dramatic advantage offered by the optics-inspired PST. 
\section{\uppercase{Response Function}}

\noindent Here we provide the underlying principle behind the superior and unique properties of the Phase Stretch Transform. The analysis will show that these properties stem from the wide dynamic range and built-in equalization inherent in the algorithm. The equations are written in two dimensions however it should be clear that they apply to one dimensional temporal waveforms or to three and higher dimensions. 

The Phase Stretch Transform (\cite{asghari2015edge}), represented by $\mathbb{S} \{ \}$, is defined by the following equation that governs the operation of PST in frequency domain to an image $E_i [x,y] $, where, x and y are two-dimensional spatial variables.

\begin{equation}\label{eq1}
  E_o [x,y]=\mathbb{S}\{E_i [x,y]\}  
\end{equation}

where the $\mathbb{S} \{ \}$  operator is defined as,
 
\begin{equation}\label{eq2}
 \mathbb{S}\big\{E_i [x,y]\big\} \triangleq IFFT2\big\{ \widetilde{K} [u,v]  \cdot \widetilde{L}[u,v] \cdot \{FFT2\{E_i [x,y]\}\big\}       
\end{equation}
 
and the complex output $E_o [x,y]$ can be defined as,
 \begin{equation}\label{eq3}
E_o [x,y]=|E_o [x,y]|  e^{j\theta[x,y]}     
\end{equation}

In the above equations,  FFT2 is the two dimensional Fast Fourier Transform, IFFT2 is the two dimensional Inverse Fast Fourier Transform and u and v are frequency variables. The function $\widetilde{K} [u,v]$ is called the warped phase kernel and the function $\widetilde{L}[u,v]$ is a localization kernel implemented in frequency domain. For simplicity, we assume here that $\widetilde{L}[u,v] =1$.

The connection with physical optics and the origin of this algorithm are as follows. Equation 3 is a two dimensional spatial extension of a one dimensional temporal optical electric field. Equation 2 describes group velocity dispersion of this light field through a medium with a dispersion induced phase $\widetilde{K} [u,v]$. PST operator is defined as the phase of the transform's output,
 \begin{equation}\label{eq4}
 PST\big\{E_i [x,y]\big\}\triangleq \measuredangle \big\{S\{E_i [x,y]\}\big\}
\end{equation}

where $\measuredangle \langle \cdot \rangle$ is the angle operator. Without the loss of generality and for simplicity, we consider operation of PST to 1D data, i.e.,
 \begin{equation}\label{eq5}
 PST\big\{E_i [x]\big\}= \measuredangle  \big\langle IFFT  \big\{\widetilde{K} [u,v]  \cdot FFT\{E_i [x]\}  \big \}  \big\rangle
\end{equation}

The warped phase kernel $\widetilde{K}[u]$ is described by a nonlinear frequency dependent phase which can be represented using taylor expansion as following 
 \begin{equation}\label{eq6}
\widetilde{K} [u] = e^{j\varphi [u]} = e^{\big ( {j \sum_{m=2}^{M}  {\dfrac{\varphi^{(m)}}{m!}} u^m }\big )}
\end{equation}
where $\varphi^{(m)}$ is the $m^{th}$-order discrete derivative of the phase $\varphi[u]$  evaluated for $u=0$ and values of m are even numbers. PST phase term  $\varphi[u]$ only contains even-order terms in its Taylor expansion due to even symmetry requirement for the phase term  $\varphi[u]$  for proper operation of PST as presented in (\cite {asghari2015edge}). In case of 2D data, we have previously used (\cite {asghari2015edge}) inverse tangent function for the phase derivative profile which leads to the following equation for PST Kernel Phase 

 \begin{equation}\label{eq7}
\begin{split}
\varphi [u,v]  =  \varphi_{polar} [r, \theta ]=  \varphi_{polar} [r] 
\\
= S \cdot  \frac{W \cdot r \cdot tan^{-1} (W.r) - ( \frac{1}{2} )  \cdot ln (1+ ( W \cdot r  ) ^{2} )  }{W \cdot  r_{max} \cdot tan^{-1} (W. r_{max}) - ( \frac{1}{2} )  \cdot ln (1+ ( W \cdot  r_{max} )^{2}  )} 
\end{split}
 \end{equation}
 
 where $ r= \sqrt{ u^{2} + v^{2} } $ and $ \theta = tan^{-1} \big( \frac{v}{u} \big) $. By controlling the PST parameters, namely,  strength S, and warp W, of the phase, edges in the image can be detected. Using the expression of warped phase kernel described in Eq. (6), output complex-field data, $E_o [x]$, can be evaluated as follows,
 \begin{equation}\label{eq8}
\begin{split}
E_o [x]= IFFT \{ {\widetilde{E}_i [u]}  \times  {\widetilde{K}[u]}  \} 
\\
= IFFT  \Big \{ {\widetilde{E}_i [u]}  \times  e^{\big ( {j \sum_{m=2}^{M}  {\dfrac{\varphi^{(m)}}{m!}} u^m }\big )} \Big \} 
\end{split}
\end{equation}     
where $\widetilde{E}_i [u]$ is the discrete Fourier transform of the input data. Simulations on images have shown that PST works best when the applied phase is small. Therefore, by restricting an applied phase that satisfies these conditions, we can use small value approximation to simplify the exponential term in Eq. (8) as
 \begin{equation}\label{eq9}
E_o [x]= IFFT \Big\{ \widetilde{E}_i [u] \times \big[ 1 +j \big( \sum_{m=2}^{M}  {\dfrac{\varphi^{(m)}}{m!}} u^m \big) \big] \Big\} 
\end{equation}  
 \begin{equation}\label{eq10}
 \rightarrow E_o [x] \approx \Big[ E_i [x] + j \sum_{m=2}^{M}  {\dfrac{ (-1)^{(m/2)}\varphi^{(m)}}{m! (2\pi)^m}} E_i [x]^{(m)} \Big]
\end{equation}  
where $E_i [x]^{(m)} $ is the $m^{th}$-order discrete derivative of the input data $E_i [x]$. As the output data is a complex quantity, the phase of the output data can be calculated as,
 \begin{equation}\label{eq11}
\begin{split}
PST\{E_i [x]\}= \measuredangle \{E_0 [x]\} 
\\
\approx \tan ^{ - 1} \Big \{ {\dfrac{{\sum_{m=2}^{M}{\dfrac{(-1)^{(m/2)}\varphi^{(m)}}{m! (2\pi)^m}}} E_i [x]^{(m)}} {E_i [x]}} \Big \}
\end{split}
\end{equation}
Finally, since the phase is restricted to small values, Eq. (11) can be simplified to,

 \begin{equation}\label{eq12}
PST\{E_i [x]\} \approx {\dfrac{{\sum_{m=2}^{M}{\dfrac{(-1)^{(m/2)}\varphi^{(m)}}{m! (2\pi)^m}}} E_i [x]^{(m)}} {E_i [x]}}
\end{equation}

The closed-form expression presented in Eq. (12) relates the PST output to the input. To give an example, the core functionality of the PST as a feature detector can be understood by closed-form expression shown in Eq. (12). The output of the PST operator is related directly to the derivatives of the input data with weighting factors of ${\dfrac{(-1)^{(m/2)}\varphi^{(m)}}{m! (2\pi)^m}}$ . Derivatives of the input data have the property to detect different features in the input data. Thus, the weighting factors can be designed to emphasize different kind of features in the input image data. In another words, PST is a reconfigurable operator that can be tuned to emphasize different features in an input image data.  

One of the crucial observations from Eq. (12) is that, output of the PST is inversely proportional to the input brightness level. Therefore, for the same contrast level change, the output is large in the dark low-light-level areas of an image. This important property, inherent in PST, equalizes the input brightness level and allows for a more sensitive feature detection and enhancement.  In the past, a lot of study has been done to improve feature detection algorithms by brightness level equalization in wide dynamic range images (\cite{hameed2011edge}). In particular, use of a Log function as a pre-processing step is one of the many commonly used techniques for brightness level equalization prior to feature detection. The log function characteristics present a higher gain for lower brightness input and vice versa. This equalizes the brightness in images and improves feature detection. Fortunately, PST operator has a built-in logarithmic behaviour in it's response function due to which it naturally works over a wide dynamic range.

\begin{figure*}[!h]
  \centering
   {\epsfig{file = 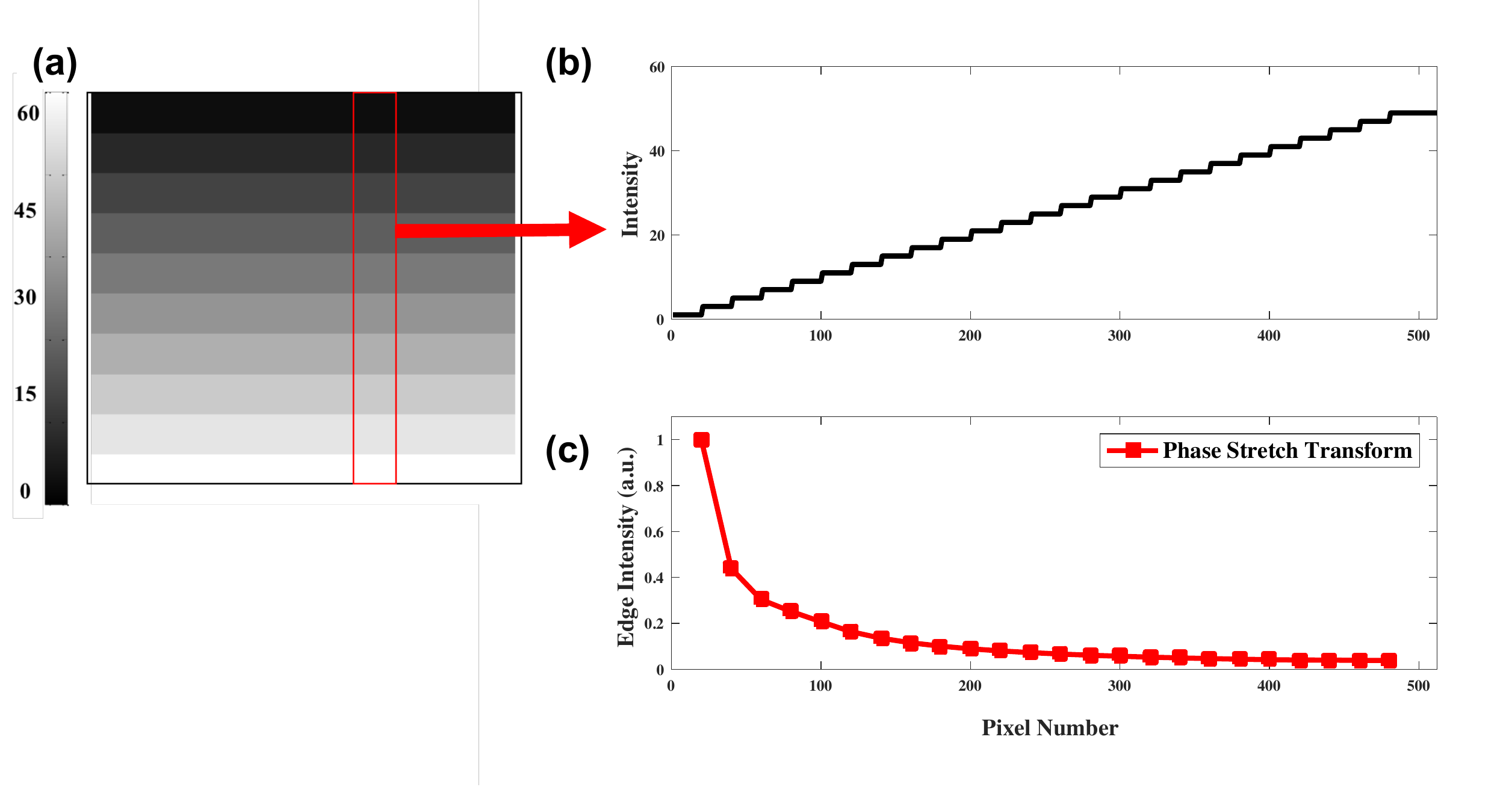, width = 16 cm}}
  \caption{Effect of Phase Stretch Transform (PST) on features with fixed contrast level change at different brightness levels.  The input data was designed to have a fixed contrast level change at different brightness levels, shown in (b). Numerically calculated output data using PST for fixed contrast level changes is different for different levels of brightness. This is due to the inverse dependence of PST output to the input brightness level described in Eq. (12).}
  \label{fig:example1}
 \end{figure*} 
 
\section{\uppercase{Simulation Results}}

In this section, we present simulation results that confirm the closed-form expression for PST derived in the previous section. We also show examples of operation of PST on digital images, supporting the new theory explained above. In the first example, we evaluate the effect of PST on features with different contrast level change at fixed brightness level and compare it to the mathematical expression derived in Eq. (12) for the PST output. The input data designed to have different contrast level change at fixed brightness level is shown in Figure 2(b). The warp, W, and strength, S, factors used for the PST operator are 12.15 and 0.48, respectively. The red-solid line representing the output data confirms that the relation of PST to  contrast level change at fixed brightness level is nonlinear. This effect is due to the brightness level equalization mechanism of PST estimated by Eq. (12). Therefore, this simulation result presented in Figure 2 confirms the accuracy of the closed-form equation to estimate the output of the PST algorithm.

Figure 3 shows another example of using PST for feature enhancement in a painting of ``Minerva of Peace''. Similar to Figure 1, the image has interesting sharp features in the scroll (see red solid box in Figure 3(a)). Results of feature detection using conventional edge derivative operator and PST operator are shown in Figure 3(b) and 3(c) respectively. Clearly, conventional edge derivative operator fails to efficiently visualize the sharp features of the alphabets in the scroll compared to the feature detection using PST as depicted in the enlarged part of the painting. However, PST traces the edges of alphabets and thus, provide more information on the contrast changes in dark areas due to its natural equalization mechanism, see Figure 3(c). Conventional edge derivative operator was implemented from find edge function in ImageJ software. The warp, W, and strength, S, factors used for the PST operator are 13 and 0.4, respectively. 

Figure 4 compare the effect of feature detection using conventional edge derivative operator with feature detection using PST on another image of planet uranus capture from a different view.The warp, W, and strength, S, factors used for the PST operator are 12.15 and 28, respectively. Conventional edge derivative operator fails to visualize the sharp contrast changes in bright areas of the image over the surface of the planet Uranus. However, PST can clearly show these surface contrast changes even in the bright areas due to its natural equalization mechanism. These surface variations over the planet are consistent with the edges detected in the Figure 1 highlighting the efficiency of PST.

 \begin{figure*}[!h]
 \centering
 {\epsfig{file = 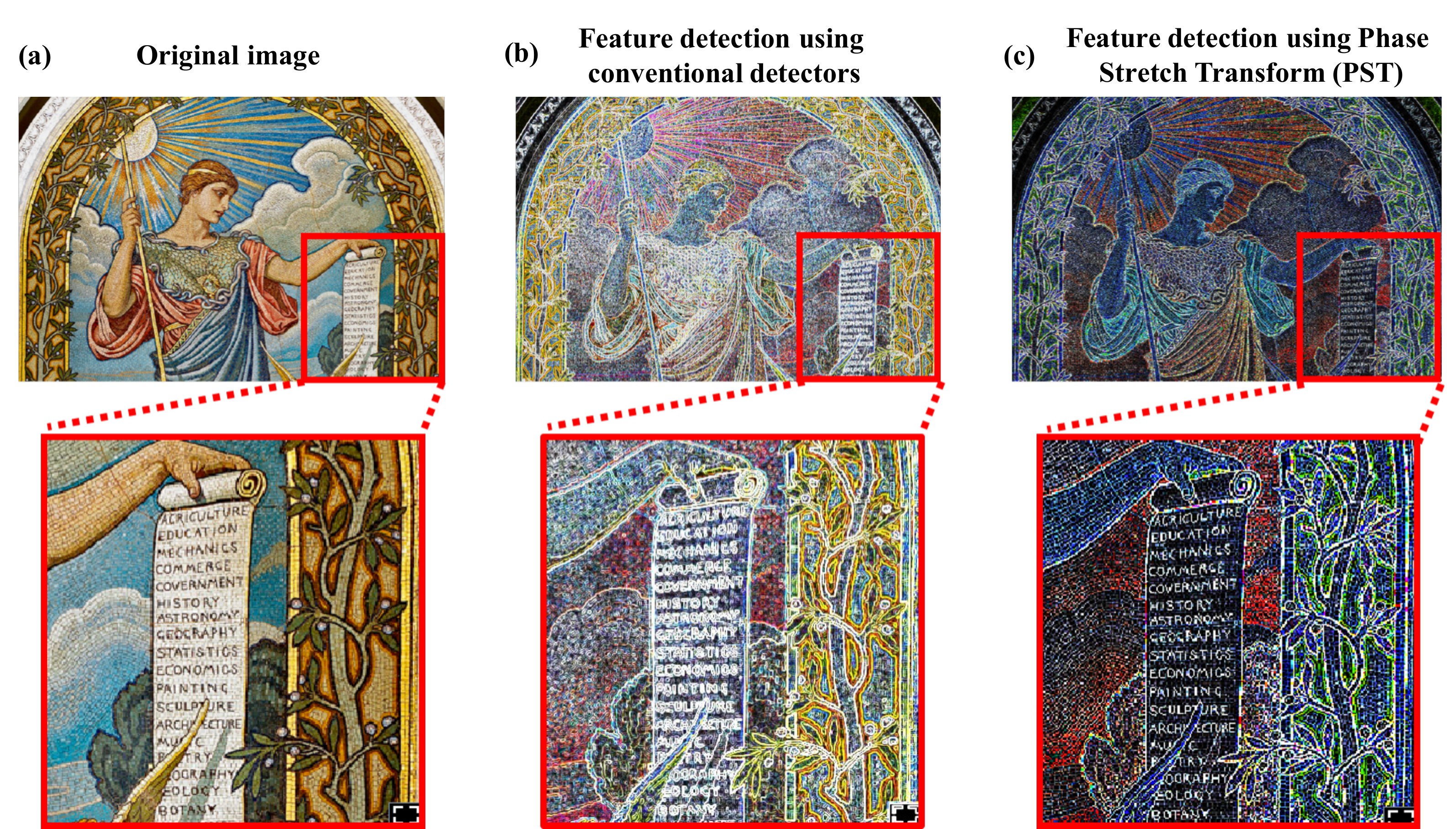, width = 16 cm}}
 \caption{Comparison of feature detection using conventional derivative operator to the case of feature detection using Phase Stretch Operator (PST). The derivative is the fundamental operation used in the popular Canny, Sobel and Prewitt edge detection methods. Original image is shown in (a). Results of feature detection using conventional edge derivative operator and PST operator are shown in (b) and (c), respectively. Enlarged view of the scroll in the painting shown in the red boxes depicts the efficiency of PST to trace the edges of alphabets in the scroll more accurately.}
 \label{fig:example1}
 \end{figure*} 

  \begin{figure*}[!h]
  \centering
   {\epsfig{file = 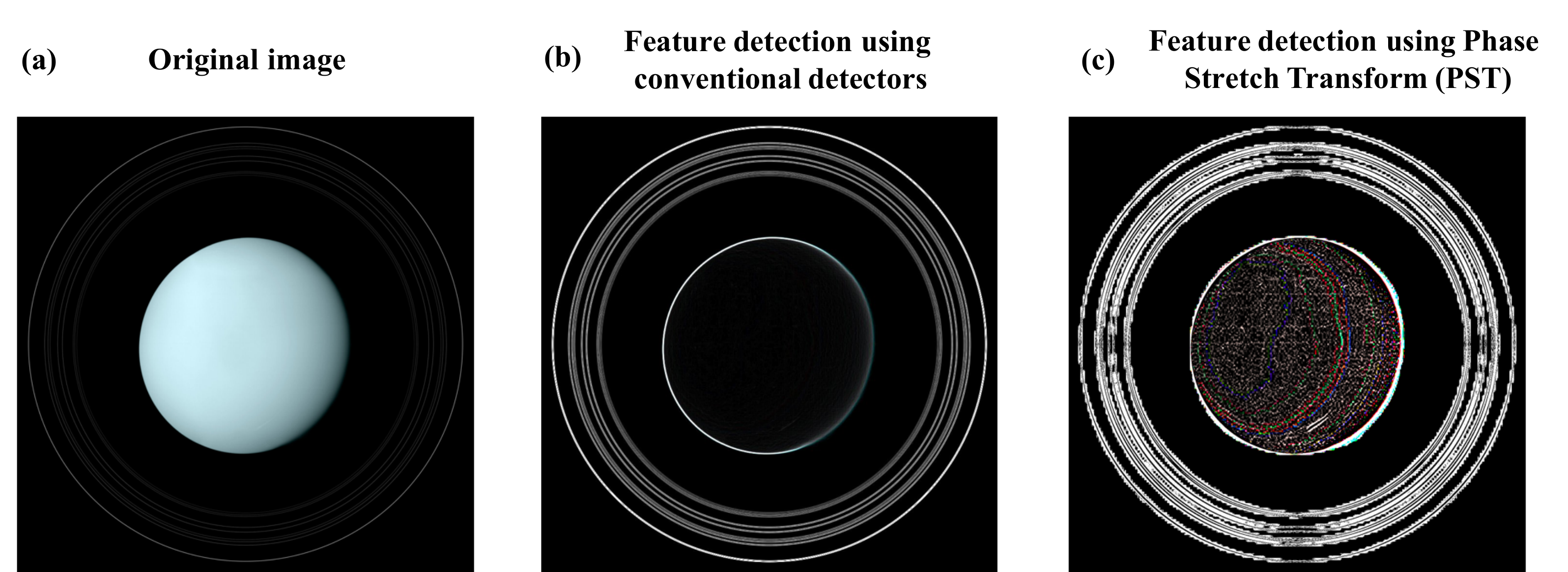, width = 16 cm}}
  \caption{Comparison of feature detection using conventional derivative operator to the case of feature detection using Phase Stretch Operator (PST) on an image of the planet Uranus captured from a different view as compared to the one shown in Figure 1. The derivative is the fundamental operation used in the popular Canny, Sobel and Prewitt edge detection methods. Original image is shown in (a). Results of feature detection using conventional edge derivative operator and PST operator are shown in (b) and (c), respectively. PST is able to locate the sharp contrast variation over the surface of the planet which are consistent with the edges located in Figure 1.}
  \label{fig:example1}
 \end{figure*} 
\section{\uppercase{Conclusions}}
New ideas are needed to deal with the exponentially increasing amount of digital data being generated. Fortunately, optics can provide a solution in certain cases. The physics of light propagation in dispersive or diffractive media has natural properties that allow it to be used for feature extraction from data. When implemented as a numerical algorithm, this concept is leading to an entirely new class of image processing techniques with high performance. 

\section*{\uppercase{Acknowledgements}}
\noindent This work was partially supported by the National Institutes of Health (NIH) grant no. 5R21 GM107924-03 and the Office of Naval Research (ONR) Multidisciplinary University Research Initiatives (MURI) program on Optical Computing.

\bibliographystyle{apalike}
{\small
\bibliography{PHOTOPTICS}}

\begin{thebibliography}{}

\bibitem[Asghari and Jalali, 2014]{asghari2014physics}
Asghari, M.~H. and Jalali, B. (2014).
\newblock Physics-inspired image edge detection.
\newblock In {\em Signal and Information Processing (GlobalSIP), 2014 IEEE
  Global Conference on}, pages 293--296. IEEE.

\bibitem[Asghari and Jalali, 2015]{asghari2015edge}
Asghari, M.~H. and Jalali, B. (2015).
\newblock Edge detection in digital images using dispersive phase stretch
  transform.
\newblock {\em Journal of Biomedical Imaging}, 2015:6.

\bibitem[{Asghari, M. H. and Jalali, B. }, 2016]{GithubURL}
{Asghari, M. H. and Jalali, B. } (2016).
\newblock {Image-feature-detection-using-Phase-Stretch-Transform}.
\newblock
  \burl{https://github.com/JalaliLabUCLA/Image-feature-detection-using-Phase-Stretch-Transform}.

\bibitem[Bhushan et~al., 1998]{bhushan1998time}
Bhushan, A., Coppinger, F., and Jalali, B. (1998).
\newblock Time-stretched analogue-to-digital conversion.
\newblock {\em Electronics Letters}, 34(11):1081--1082.

\bibitem[Chen et~al., 2016]{chen2016deep}
Chen, C.~L., Mahjoubfar, A., Tai, L.-C., Blaby, I.~K., Huang, A., Niazi, K.~R.,
  and Jalali, B. (2016).
\newblock Deep learning in label-free cell classification.
\newblock {\em Scientific reports}, 6.

\bibitem[Hameed and Wang, 2011]{hameed2011edge}
Hameed, Z. and Wang, C. (2011).
\newblock Edge detection using histogram equalization and multi-filtering
  process.
\newblock In {\em 2011 IEEE International Symposium of Circuits and Systems
  (ISCAS)}, pages 1077--1080. IEEE.

\bibitem[Han and Jalali, 2003]{han2003photonic}
Han, Y. and Jalali, B. (2003).
\newblock Photonic time-stretched analog-to-digital converter: fundamental
  concepts and practical considerations.
\newblock {\em Journal of Lightwave Technology}, 21(12):3085.

\bibitem[Herink et~al., 2016]{herink2016resolving}
Herink, G., Jalali, B., Ropers, C., and Solli, D. (2016).
\newblock Resolving the build-up of femtosecond mode-locking with single-shot
  spectroscopy at 90 mhz frame rate.
\newblock {\em Nature Photonics}.

\bibitem[{IBM}, 2016]{IBMURL}
{IBM} (2016).
\newblock {Bringing big data to the enterprise}.
\newblock
  \burl{https://www-01.ibm.com/software/data/bigdata/what-is-big-data.html}.

\bibitem[Ilioudis et~al., 2015]{ilioudis2015edge}
Ilioudis, C.~V., Clemente, C., Asghari, M.~H., Jalali, B., and Soraghan, J.~J.
  (2015).
\newblock Edge detection in sar images using phase stretch transform.

\bibitem[Ilovitsh et~al., 2016]{ilovitsh2016phase}
Ilovitsh, T., Jalali, B., Asghari, M.~H., and Zalevsky, Z. (2016).
\newblock Phase stretch transform for super-resolution localization microscopy.
\newblock {\em Biomedical Optics Express}, 7(10):4198--4209.

\bibitem[Jalali and Mahjoubfar, 2015]{jalali2015tailoring}
Jalali, B. and Mahjoubfar, A. (2015).
\newblock Tailoring wideband signals with a photonic hardware accelerator.
\newblock {\em Proceedings of the IEEE}, 103(7):1071--1086.

\bibitem[Mahjoubfar et~al., 2015]{mahjoubfar2015design}
Mahjoubfar, A., Chen, C.~L., and Jalali, B. (2015).
\newblock Design of warped stretch transform.
\newblock {\em Scientific reports}, 5.

\bibitem[Ng et~al., 2014]{ng2014demonstration}
Ng, W., Rockwood, T., and Reamon, A. (2014).
\newblock Demonstration of channel-stitched photonic time-stretch
  analog-to-digital converter with enob≥ 8 for a 10 ghz signal bandwidth.
\newblock In {\em Proceedings of the Government Microcircuit Applications \&
  Critical Technology Conference (GOMACTech'14)}.

\bibitem[Roussel et~al., 2014]{roussel2014microbunching}
Roussel, E., Evain, C., Szwaj, C., Bielawski, S., Raasch, J., Thoma, P.,
  Scheuring, A., Hofherr, M., Ilin, K., W{\"u}nsch, S., et~al. (2014).
\newblock Microbunching instability in relativistic electron bunches: Direct
  observations of the microstructures using ultrafast ybco detectors.
\newblock {\em Physical review letters}, 113(9):094801.

\bibitem[Solli et~al., 2007]{solli2007optical}
Solli, D., Ropers, C., Koonath, P., and Jalali, B. (2007).
\newblock Optical rogue waves.
\newblock {\em Nature}, 450(7172):1054--1057.

\bibitem[Suthar, 2016]{suthar2016decision}
Suthar, M. (2016).
\newblock Decision support systems for radiologists based on phase stretch
  transform.

\bibitem[Suthar et~al., 2016]{suthar2016diagnostic}
Suthar, M., Mahjoubfar, A., Seals, K., Lee, E.~W., and Jalaii, B. (2016).
\newblock Diagnostic tool for pneumothorax.
\newblock In {\em Photonics Society Summer Topical Meeting Series (SUM), 2016
  IEEE}, pages 218--219. IEEE.

\end{thebibliography}

\vfill
\end{document}